# IFCNet: A Benchmark Dataset for IFC Entity Classification

Christoph Emunds, Nicolas Pauen, Veronika Richter, Jérôme Frisch, Christoph van Treeck
Institute of Energy Efficiency and Sustainable Building E3D, RWTH Aachen University, Germany
emunds@e3d.rwth-aachen.de

**Abstract.** Enhancing interoperability and information exchange between domain-specific software products for BIM is an important aspect in the Architecture, Engineering, Construction and Operations industry. Recent research started investigating methods from the areas of machine and deep learning for semantic enrichment of BIM models. However, training and evaluation of these machine learning algorithms requires sufficiently large and comprehensive datasets. This work presents IFCNet, a dataset of single-entity IFC files spanning a broad range of IFC classes containing both geometric and semantic information. Using only the geometric information of objects, the experiments show that three different deep learning models are able to achieve good classification performance.

## 1. Introduction

Enhancing interoperability between domain-specific information modeling processes and, thus, software products for Building Information Modeling (BIM) is an important aspect to improve the lifecycle support of buildings and to facilitate the collaboration of the different disciplines across Architecture, Engineering, Construction and Operations (AECO). The Industry Foundation Classes (IFC) provide an open data exchange format for sharing information between these stakeholders.

However, since the IFC standard has to cover a broad spectrum of concepts, it contains a large number of entities and is highly complex. Past studies have shown that IFC-based exchanges of models are prone to an information loss due to reduction, simplification or interpretation when sharing data between multiple specialized software products (Bazjanac & Kiviniemi, 2007). One major issue is a potential mismapping between native BIM elements and IFC entities, which can arise through e.g. manual error during model creation or the reliance on default templates (Belsky, et al., 2016). Furthermore, CAD software products interpret specifications differently when processing in- and output data.

When sharing BIM models with other teams, semantic integrity is a prerequisite for a seamless workflow and effective collaboration. Many specialized applications rely on accurate semantic information to perform their tasks, e.g. energy efficiency modelling (Schlueter & Thesseling, 2009; Ham & Golparvar-Fard, 2015) or code compliance checking (Eastman, et al., 2009). Inconsistent object classification has been identified to be a common interoperability issue between different BIM authoring software suites (Belsky et al., 2016; Lai & Deng, 2018).

Researchers have started approaching this issue with methods from the area of machine and deep learning (Bloch & Sacks, 2018). These algorithms typically need labelled datasets to learn from. However, comprehensive and rich datasets in the domain of BIM and IFC are scarce, which makes the development and verification of such models difficult. In this work, the authors introduce a benchmark dataset of single-entity IFC files covering a broad range of IFC classes. This dataset, named *IFCNet[1]*, should contribute to the standardization of performance evaluations of future work in this domain. To evaluate the usefulness of IFCNet, three deep learning methods are trained to classify the entities and their performance is reviewed.

---
[1] https://ifcnet.e3d.rwth-aachen.de

The key contributions of this research paper can be summarized as follows:

- A benchmark dataset for IFC entity classification, named *IFCNet*, is released.
- The application of recent advances in the area of geometric deep learning to the classification of IFC elements is shown using three different approaches.
- An evaluation of these deep learning methods is conducted to demonstrate the opportunities and challenges posed by IFCNet.

## 2. Related Work

Existing approaches for the classification of BIM and IFC elements can be categorized into rule-based and machine-learning-based methods. Thomson and Boehm (2015) use RANSAC to identify dominant planes and reconstruct IFC geometry from 3D point clouds, followed by an optional step of geometric reasoning. Others have used region growing (Dimitrov & Golparvar-Fard, 2015) or surface normal approaches (Barnea & Filin, 2013). Sacks et al. (2017) derived rules for object classification using object features and spatial relationships between object pairs. Ma et al. (2017) devise a semantic enrichment process by establishing a knowledge base that associates objects via their geometric and spatial features.

While these methods proof to work well on specific cases, Bloch & Sacks (2018) argue that rule-based workflows are not applicable to all problems. In recent work, researchers started exploring algorithms from the areas of machine and deep learning. Koo et al. (2020) apply PointNet (Qi, et al., 2017) and a Multi-view Convolutional Neural Network (MVCNN) (Su, et al., 2015) to classify elements of road infrastructure. Kim et al. (2019) use images of objects to train a 2D CNN to recognize furniture elements. Leonhardt et al. (2020) also employ PointNet for classification of IFC objects and for semantic segmentation of rooms.

Many of these works assemble their own datasets, but do not release them publicly. On one hand, this is inefficient, since these datasets cannot be used by the research community and thus work is done repeatedly. On the other hand, it makes comparisons between different methods impossible. IFCNet's goal is to serve as a benchmark to be used by other researchers to develop, train, and test their methods and algorithms on and offer a common ground for comparing them.

## 3. The IFCNet Dataset

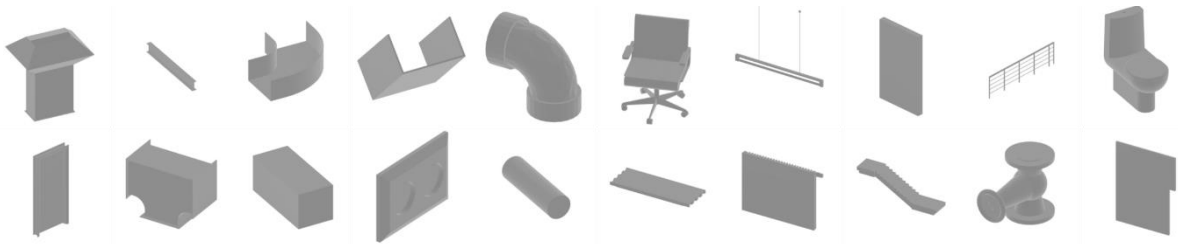

Figure 1: Example objects for each of the 20 classes of IFCNetCore.

To assemble IFCNet, around 1000 IFC models were collected from real-world projects, student works and online sources, such as the open IFC model repository of the university of Auckland (Dimyadi, et al., 2010). The models were created with different authoring software products, most notably Autodesk Revit and ArchiCAD. Afterwards, the models were decomposed into individual entities by extracting all objects into separate files. For the first version of IFCNet, the focus has been put on the subtypes of IfcDistributionElement, IfcBuildingElement and IfcFurnishingElement. Additionally, the attached IfcPropertySets have been extracted as well. The extraction results in roughly 1.2M entities from 82 different IFC classes. The data contain several different representation types, including Brep, AdvancedBrep, MappedRepresentation, SweptSolid, Tesselation and CSG.

The resulting IFC files are deduplicated to eliminate objects with identical geometry. To be able to perform this deduplication in linear time, the vertices of every object are normalized to the unit sphere and used as the key in a hash map. Objects with identical sets of vertices are then mapped onto the same key and can thus be eliminated. This, of course, assumes that the vertices match exactly, meaning that the vertices of two different objects which are in fact duplicates need to be in the same order. However, this was found to be true for the majority of objects, judging by the fact that this naïve way of deduplicating geometries reduces the aforementioned 1.2M to around 290k entities. Since objects are only deduplicated within their respective class, this process can easily be parallelized.

In the next step, the entities are reviewed manually and misclassifications are corrected. To support this process, a web-based tool was developed, which allows users to review an entity's geometric representation and attached metadata before confirming or changing its class and enables quick switching between the different IFC classes and their objects. Furthermore, the tool supports exploring the already labelled data to document the current progress of the dataset. The labelling process has been carried out and supervised by domain experts to ensure the quality of the dataset. A view of this tool is shown in Figure 2, displaying an IfcValve.

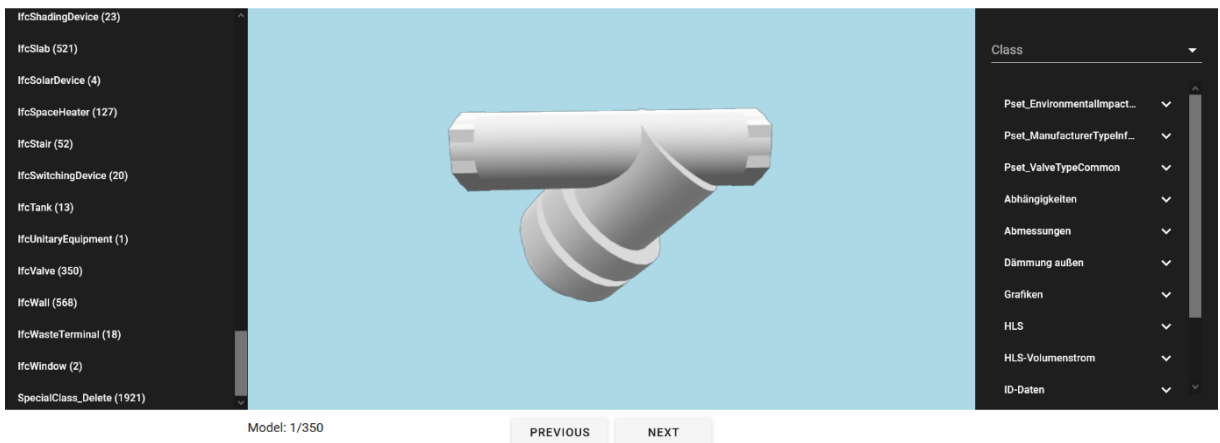

Figure 2: View of the tool used during the labelling process. The menu on the left allows switching between IFC classes. The menu on the right displays the current object's properties. The central canvas shows objects of the selected class in 3D.

The full IFCNet dataset currently consists of 19,613 confirmed objects distributed over 65 classes, most of which are highly imbalanced with respect to the number of objects they contain. Therefore, a subset of 20 classes is selected for the experiments in Section 4. The first version of this sub-dataset, called *IFCNetCore*, contains a total of 7,930 objects (Figure 1). Table 1 shows the number of objects per class before and after deduplication of the full dataset, as well as the training and test split for IFCNetCore.

Some classes, like IfcWall, have little intra-class variance, while others have very large intra-class variance. IfcFurniture for example contains vastly different types of furniture, from chairs and tables to wardrobes. An additional challenge is posed by a small inter-class variance between certain classes like IfcWall and IfcPlate, both of which are rectangular shapes with varying thickness and little to no details with respect to their geometry. To better reflect the reality of people working with IFC, models and elements of different Level of Information Need (LOIN) have been included, which also covers objects that have a placeholder appearance (e.g. generic-looking cubes) and are thus likely to only be classifiable through their metadata.

Most IFC objects have additional metadata in the form of IfcProperties, which are grouped together via IfcPropertySets. The simplest and most frequently used kind of properties are user-defined key-value pairs, which often come in different languages. For instance, German, English, Dutch and French have been observed throughout the labelling process.

Table 1: Number of objects per class in IFCNet and IFCNetCore. A class in IFCNetCore can have more objects than there were after deduplication, since e.g. IfcBuildingElementProxy objects could have moved into that class during the labelling process. Note that not all of the objects listed under *after deduplication* have been reviewed and confirmed, yet.

| Class | Before deduplication | After deduplication | Training set | Test set |
|---|---|---|---|---|
| IfcAirTerminal | 6,227 | 496 | 333 | 142 |
| IfcBeam | 128,027 | 17,957 | 198 | 84 |
| IfcCableCarrierFitting | 2,913 | 511 | 361 | 155 |
| IfcCableCarrierSegment | 4,135 | 2,820 | 370 | 159 |
| IfcDoor | 11,569 | 1,833 | 216 | 93 |
| IfcDuctFitting | 29,409 | 7,590 | 455 | 195 |
| IfcDuctSegment | 28,783 | 22,129 | 372 | 159 |
| IfcFurniture | 7,943 | 218 | 157 | 67 |
| IfcLamp | 0 | 0 | 65 | 27 |
| IfcOutlet | 1,559 | 34 | 41 | 18 |
| IfcPipeFitting | 117,979 | 5,510 | 454 | 194 |
| IfcPipeSegment | 170,308 | 86,979 | 454 | 195 |
| IfcPlate | 20,472 | 3,436 | 366 | 157 |
| IfcRailing | 3,112 | 967 | 295 | 127 |
| IfcSanitaryTerminal | 289 | 32 | 316 | 136 |
| IfcSlab | 8,787 | 4,018 | 355 | 152 |
| IfcSpaceHeater | 392 | 51 | 89 | 38 |
| IfcStair | 807 | 80 | 36 | 16 |
| IfcValve | 15,335 | 362 | 242 | 104 |
| IfcWall | 21,336 | 8,348 | 376 | 161 |
| Total | | | 5551 | 2379 |

## 4. Experiments

The following experiments apply three neural network approaches to the IFCNetCore dataset. These architectures were chosen because they are among the current state-of-the-art and cover a broad range of intuitive representations for 3D data, i.e. 2D projections, point clouds and triangulated meshes. However, all of these methods only consider the objects' geometric information. Investigating ways to combine geometric and semantic information during training is beyond the scope of this paper. The code for the neural network models is based on the PyTorch implementations of the original publications.

All experiments follow the same training protocol: The IFCNetCore dataset is split into a training and a test set. Afterwards, the data is transformed into the format expected by the different architectures. To determine the best set of hyperparameters, 30% of the training data is split off into a validation set. The models are then trained on the remaining 70% of the training data and evaluated on the validation set after each epoch. The balanced accuracy metric is used to decide for the best performing configuration of hyperparameters. Finally, the models are trained once more on the whole training set with fixed hyperparameters. Evaluation on the test set only occurs once at the end of this procedure. The code used to conduct these experiments will be released along with this work[2].

### 4.1 MVCNN

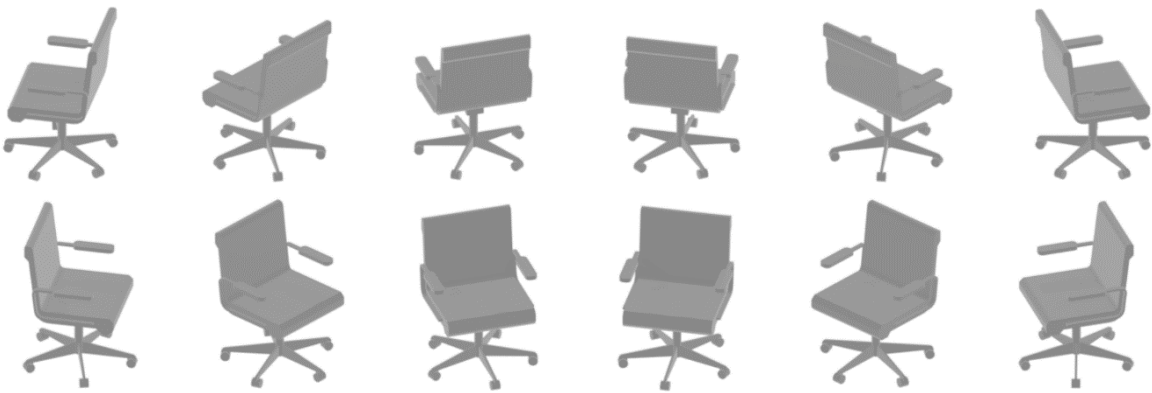

Figure 3:   Example of a set of 12 views to be consumed by the MVCNN

The Multi-View Convolutional Neural Network (MVCNN) combines information from multiple views of a 3D shape to learn a shape descriptor (Su, et al., 2015). Since MVCNN uses rendered 2D views of an object from several perspectives, it has two advantages over the other methods presented here. First, neural networks for image classification have received much more attention in Deep Learning research over the last years. Neural network building blocks like 2D convolutions have been specifically designed to work well on image data. Second, MVCNN benefits from the existence of other large-scale image datasets like ImageNet (Deng, et al., 2009). CNN architectures are commonly pre-trained on these massive datasets and can later be fine-tuned on much smaller datasets while still performing well.

To prepare the IFCNetCore dataset to be consumed by MVCNN, 12 views are rendered for each object by a camera rotating around the object's up-axis in 30° increments (Figure 3). Similar to Su et al. (2015), the Phong reflection model (Phong, 1975) is used to generate the rendered views. Figure 4 shows the results of the evaluation on the test set.

---
[2] https://github.com/cemunds/ifcnet-models

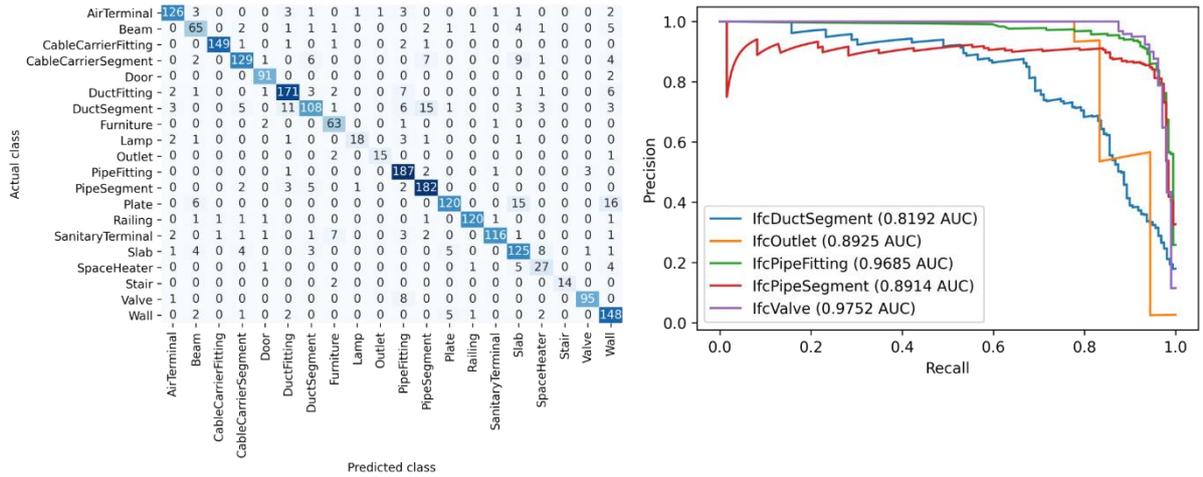

Figure 4 **Left**: Confusion matrix of the MVCNN model. **Right**: Precision-recall curves for selected IFC classes with corresponding values for Area Under the Curve (AUC).

## 4.2 DGCNN

The Dynamic Graph Convolutional Neural Network (DGCNN) (Wang, et al., 2019) is inspired by PointNet (Qi, et al., 2017), but operates on neighborhoods of points by using convolution operations. This allows DGCNN to exploit local geometric structures.

During pre-processing, 2048 points are sampled uniformly at random from each object in IFCNetCore. The point clouds are normalized to the unit sphere before they are fed through the model. The results of the evaluation on the test set are shown in Figure 5.

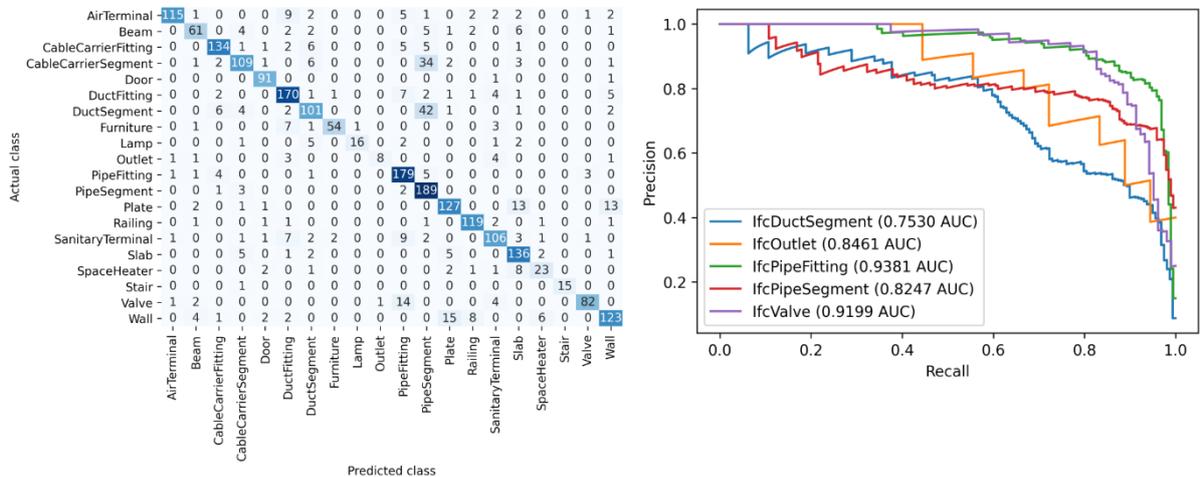

Figure 5 **Left**: Confusion matrix of the DGCNN model. **Right**: Precision-recall curves for selected IFC entities with corresponding values for Area Under the Curve (AUC).

## 4.3 MeshNet

In contrast to the previous two methods, MeshNet (Feng, et al., 2018) uses the geometric information of the mesh directly to learn a classifier. It solves the complexity and irregularity problem of mesh data by regarding the faces as the unit and using per-face processes and a symmetry function. Moreover, it splits faces into spatial and structural features by using a spatial and structural descriptor and a mesh convolution block.

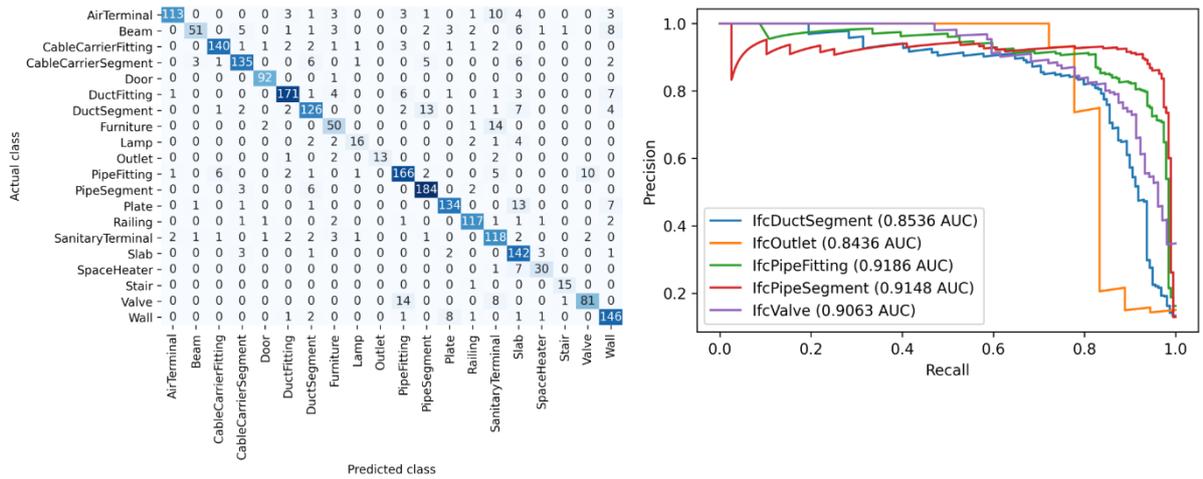

Figure 6 **Left**: Confusion matrix of the MeshNet model. **Right**: Precision-recall curves for selected IFC entities with corresponding values for Area Under the Curve (AUC).

Before training, the meshes are simplified to a maximum of 2048 faces using MeshLab's (Cignoni, et al., 2008) implementation of Quadric Edge Collapse Decimation. Afterwards, the meshes are converted into lists of faces containing information about their center, corners and normal as well as their immediate neighboring faces. The results on the test set are shown in Figure 6.

### 4.4 Comparison

A comparison of precision-recall curves for selected classes can be seen in Figure 7. Not surprisingly, classes for which there are very few objects like IfcOutlet show a worse performance. However, MVCNN still achieves better results than DGCNN and MeshNet.

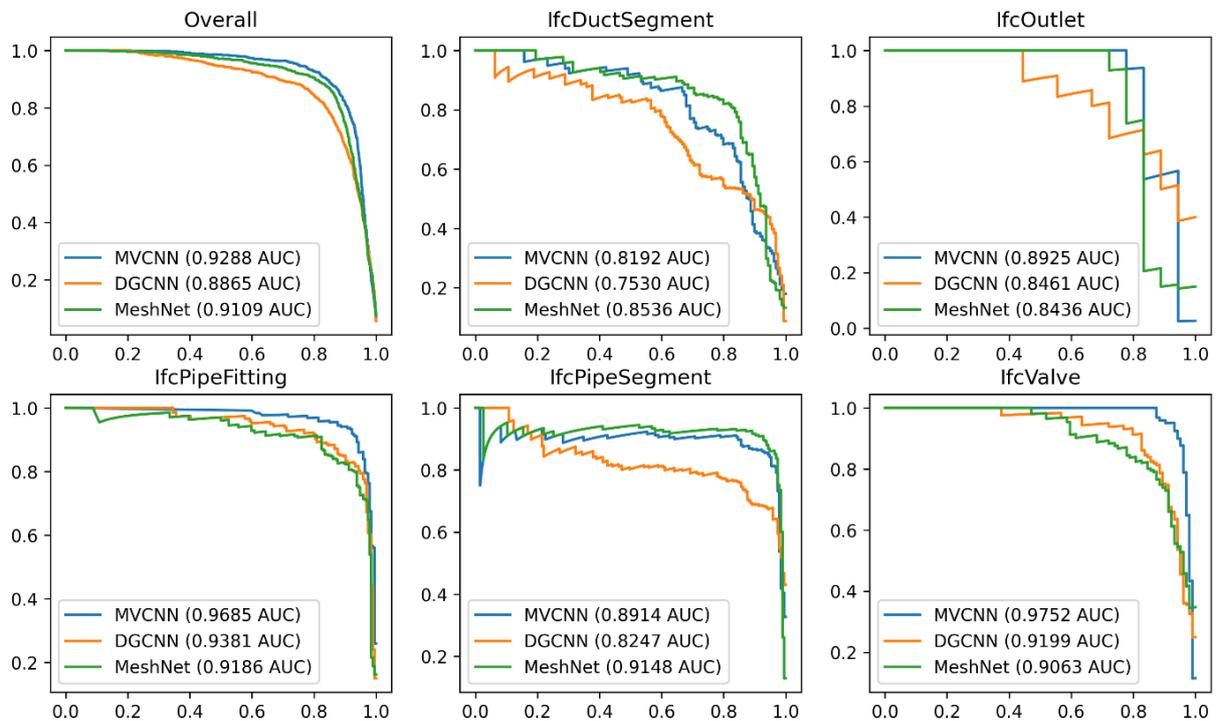

Figure 7: Comparison of precision-recall curves for selected classes. The top-left plot shows the micro-averaged precision-recall curve over all classes.

Table 2: Results of the evaluation on the test set for the three models.

| Model | Balanced Accuracy | F1 score |
|---|---|---|
| MVCNN | **85.54%** | **86.93%** |
| DGCNN | 79.11% | 82.15% |
| MeshNet | 83.32% | 85.72% |

Table 2 shows the balanced accuracy and F1 score for the three models. MVCNN achieves the best overall results. The confusion matrices show that each of the three models has its own strengths and weaknesses, but that they also make similar mistakes. For instance, plates, slabs and walls are among the most confused classes. Another example of commonly confused classes are duct segments and pipe segments. However, all three models show a reasonable performance and proof that they are able to learn from the dataset.

## 5. Limitations

Notably, the absolute sizes of objects are lost due to normalization of the data before training. Incorporating this information into the classification process is likely to improve results for objects that might look similar, but differ greatly in size. Moreover, some objects, especially those with few geometric details, might be very hard to classify when taken out of the context of the full BIM model or without regarding their semantic information. However, most current neural network architectures are trained end-to-end from raw data and were not designed to consume such explicit features.

With such a large quantity of objects, it is difficult to ensure uniqueness. The deduplication process is able to eliminate objects with identical geometry. However, there are many objects that look alike to a human observer, but are not identical on the mesh level. Such cases include permutations of a mesh's vertices or non-uniform scaling along axes. To conduct an exhaustive search and also detect objects that are *almost* identical is not feasible. Further research will have to investigate more efficient methods to eliminate near-duplicate objects.

Many objects use a mix of languages in their metadata. Sometimes the value of a key-value pair might be missing if a field in the authoring software was left unset. Moreover, it is common to encounter abbreviations and acronyms, which require a certain amount of domain knowledge to make use of. In some cases, properties might also be inaccurate or simply wrong. These issues make it difficult to incorporate the semantic properties into the classification process without thorough pre-processing.

## 6. Conclusion and Future Work

The first version of IFCNetCore offers a common benchmark for model training and evaluation. Expanding IFCNet with more objects, classes, and semantic information is an ongoing effort to create a large-scale dataset for the BIM and IFC domain. Since the labelling process requires specific domain knowledge, creating this dataset is even more resource intensive than other datasets used in machine and deep learning. The goal for IFCNet is to become a useful resource for other researchers working on semantic enrichment of BIM models.

The results of the experiments conducted on IFCNetCore show a good classification performance, despite only using the geometric information of objects. Further research could investigate models that can take the properties and semantic information into account to improve on these results. Moreover, in the domain of 2D images, models used for segmentation or detection are commonly pre-trained on large-scale image datasets. How to effectively use such transfer learning approaches for 3D data is an area of active research. One could imagine that, with sufficient size of IFCNet, it should be possible to use the dataset for similar pre-training purposes.

The classification process presented in this work can be integrated into the BIM workflow similarly to the SEEBIM method of Belsky et al. (2016). Upon import of an IFC file into a BIM tool, the trained network is used to infer the classes of the individual elements of the model. Afterwards, the author is prompted with a screen showing the potential misclassifications and can then decide to accept or reject the propositions of the network.

**Acknowledgements**

The research within the project EnergyTWIN leading to these results has received funding from the German Ministry for Industry and Energy under grant agreement no. 03EN1026A. The authors would like to thank the Geodetic Institute and Chair for Computing in Civil Engineering & Geo Information Systems at RWTH Aachen, aedifion GmbH, DiConneX GmbH, TEMA Technologie Marketing AG, Internet Marketing Services GmbH and Aachener Grundvermögen Kapitalverwaltungsgesellschaft mbH for their contribution to the project.